\documentclass{clv3} 
\usepackage{pifont}
\usepackage{enumitem}
\usepackage{xspace}
\usepackage{booktabs}
\usepackage{fancyvrb}

\usepackage{amsmath,amsfonts,bm}

\newcommand{\fromto}{\rightarrow}

\def\secref#1{Section \ref{#1}}

\def\eqref#1{equation~\ref{#1}}

\def\1{\bm{1}}

\def\vx{{\bm{x}}}

\DeclareMathAlphabet{\mathsfit}{\encodingdefault}{\sfdefault}{m}{sl}
\SetMathAlphabet{\mathsfit}{bold}{\encodingdefault}{\sfdefault}{bx}{n}

\def\gI{{\mathcal{I}}}

\def\gX{{\mathcal{X}}}

\usepackage{tikz}
\let\numdefbak\numdef{}
\let\endnumdefbak\endnumdef{}
\let\numdef\relax
\let\endnumdef\relax
\usepackage{hyperref}
\let\numdef\numdefbak{}
\let\endnumdef\endnumdefbak{}
\usepackage{url}
\usepackage[textsize=scriptsize]{todonotes}
\setuptodonotes{fancyline, color=yellow!20}

\newcommand{\changetk}[1]{#1} 

\newcommand*\samethanks[1][\value{footnote}]{\hypersetup{linkcolor=blue}\footnotemark[#1]}

\title{Problem Solving Through Human-AI Preference-Based Cooperation}

\author{Subhabrata Dutta}
\affil{TU Darmstadt}
\author{Timo Kaufman}
\affil{LMU Munich, MCML}
\author{Goran Glava\v{s}}
\affil{University of W\"{u}rzburg}
\author{Ivan Habernal}
\affil{RU Bochum}
\author{Kristian Kersting}
\affil{TU Darmstadt}
\author{Frauke Kreuter}
\affil{LMU Munich, MCML}
\author{Mira Mezini}
\affil{TU Darmstadt}
\author{Iryna Gurevych}
\affil{TU Darmstadt}
\author{Eyke H\"{u}llermeier\thanks{Shared senior authorship}}
\affil{LMU Munich, MCML}
\author{Hinrich Sch\"{u}tze\samethanks}
\affil{LMU Munich, MCML}

\def\month{MM}  
\def\year{YYYY} 
\newcommand{\framework}{{\tt HAI-Co\textsuperscript{2}}\xspace} 
\newcommand{\fullname}{{{\bf H}uman-{\bf AI} {\bf Co-Co}nstruction}\xspace}

\newcounter{notecounter}

\newcommand{\enoteson}{\long\gdef\enote##1##2{{
\stepcounter{notecounter}
{\large\bf
\hspace{0cm}\arabic{notecounter} $<<<$ ##1: ##2
$>>>$\hspace{1cm}}}}}

\enoteson{}

\usepackage[capitalize]{cleveref}

\begin{document}
\dochead{Position paper}

\runningtitle{Problem Solving Through Human-AI Preference-Based Cooperation}
\runningauthor{Dutta et al.}
\maketitle

\begin{abstract}
While there is a widespread belief that artificial general
intelligence (AGI) -- or even superhuman AI -- is imminent,
complex problems in expert domains are far from being
solved. We argue that such problems require human-AI
cooperation and that the current state of the art in
generative AI is unable to play the role of a reliable
partner due to a multitude of shortcomings, including
difficulty to keep track of a complex solution artifact
(e.g., a software program), limited support for versatile
human preference expression and lack of adapting to human
preference in an interactive setting. To address these
challenges, we propose \framework{}, a novel human-AI
co-construction framework. We take first steps towards a
formalization of \framework{} and
discuss the difficult open research problems that it faces.
\end{abstract}


\section{Introduction}
\label{sec:intro}
Despite the impressive advances of generative AI
\changetk{\citep{cao2025survey}}, especially for natural
language (large language models), vision (vision language
models) and code (code models), recent investigations have
pointed out a lack of competence in dealing with complex
generation problems that require intricate
planning~\citep{kambhampati2024llms} and task adherence while
keeping track of multiple
constraints~\citep{xie2024travelplanner}. A broad class of
such complex problems, especially complex problems in expert
domains, requires active human participation. Therefore,
although the recent focus in generative AI has mostly been
on complete automation
\citep{hong2024metagpt,brown2020language}, we believe that
human-AI cooperation is a more promising approach.



To address complex problems of this kind, we propose
\fullname{} (\framework{}), a novel framework for human-AI
cooperative problem solving that builds on preference-based
learning and search methodology and relies on natural
language to facilitate interaction.


{

We acknowledge that many substrands of the NLP, ML, AI and
HCI communities have noticed and addressed problems that are
closely related to the problem \framework is addressing.
Moreover, many of the components of \framework are also
components of this prior work. The differentiator of our
proposal lies (i) in the specifics of the problem statement,
(ii) in the specifics of the holistic framework that we
call \framework and (iii) in a clear articulation of the
research challenges and unmet needs that follow from the way
we define problem and framework.  We will lay out the
research challenges in \secref{sec:challenges} and review
related work in detail in \secref{sec:rw}. To set the
context for the reader, we highlight here the most
important points that make our work novel.

First, we address problems in \textbf{expert domains that require
complex solutions or artifacts}.
\begin{itemize}
\item We posit that an explicit and persistent representation of the
solution
space is crucial for systematic solution construction and
that this solution space must be equipped to represent the
complexities of expert domains, including through an
abstraction hierarchy.
\item We propose search as the paradigm for constructing
solutions: construction proceeds from an
initial draft to a satisfactory solution step by step where
each step consists of a search for an appropriate extension
or modification of the current artifact.
\end{itemize}

Second, \framework
is designed for \textbf{complex problem solving by a team of an
expert and an AI agent}.
\begin{itemize}
\item In contrast to many other approaches, \framework
is set up for a cooperation of expert and agent as equal
partners, each contributing their complementary strengths.
\item In complex domains, the exact goal of the cooperation
is often underspecified in the beginning. This means that
the cooperation is not only about constructing the solution,
but also about constructing the precise objective of the solution. This raises
ethical concerns (e.g., influencable reward functions) that
need to be addressed.
\item We posit natural language as the primary medium
of communication. One of the challenges in \framework is
that the agent has to learn effectively from implicit
human feedback in natural language, but also from other signals in the complex
co-construction environment, e.g., multimodal information
and human edits. 
\item Finally, a new evaluation methodology needs to be
developed for \framework due to
the difficulty of assessing the quality of solutions in
complex expert domains and
due to the
open-endedness  and non-uniqueness of solutions to complex
problems in expert domains.
\end{itemize}

}

In this article, we
first give a general introduction to \framework{}
(\secref{sec:nonformalframework}), including its ethical
challenges, and
take first steps towards 
 a formalization
(\secref{sec:framework}).  We outline open research
challenges posed by \framework in \secref{sec:challenges}.
\secref{sec:rw} discusses related work.
\secref{sec:conclusion} concludes the paper.

\section{\framework{}: Human-AI co-construction through preference-based search}
\label{sec:nonformalframework}
In this article,
we propose
\fullname{} (\framework{}),
a novel framework for human-AI
cooperative problem solving.
The four defining characteristics of \framework{} are (i)
solution of complex problems in expert domains that require active human
participation, (ii)
co-construction of the solution by human and AI agent,
(iii) co-construction of the objective by human and AI agent by
means of preference learning and (iv) the use of natural
language as the main communication medium, which makes it
possible for human and AI agent to be equal partners, with
complementary strengths, in the co-construction. We now
describe these four characteristics in more detail.

First, we target
applications in \textbf{expert domains} where the task is
to construct a solution to a \textbf{complex problem}.
Since expert domains are our focus, we use
``expert'' and ``human'' interchangeably in this article.
Examples of complex problems in expert domains include
writing a computer program in software engineering;
constructing a machine
learning pipeline in automated machine learning (AutoML);
writing a related work section in scientific research;
and developing a formalization of a problem described in
natural language in mathematics.

Second, we conceive of problem solving
as
\textbf{co-construction of the solution to the complex problem by a
human and an AI agent} or -- more generally -- by a team of
humans and agents.
As detailed in \secref{sec:framework},
the problem solving process is
formalized as a process of \emph{systematic search} in a
\emph{construction space} $\mathcal{X}$ of \emph{candidate
solutions} on several hierarchical levels of abstraction.
In this
co-constructive process,  candidate solutions
are modified step by step until a sufficiently
good solution has been found.

We draw inspiration from our understanding of how humans
collectively devise solutions to complex problems.  Humans
often tackle such problems by iteratively co-constructing a
solution step by step, revising and refining draft solutions
while transitioning between different levels of abstraction
and exchanging information about preferences and potential
improvements in natural language.  Our primary
motivation for \framework comes from natural language
processing and computer science;
see \secref{sec:rw}
for a brief discussion
of related
fields that
have conducted extensive research on human-human cooperation
on solving tasks.

Human-human cooperation would make less sense as a promising template
for human-AI cooperation if current AI systems could solve
complex expert-domain problems  on their own.
However, current AI
capabilities are limited
for complex expert-domain problems,
e.g.,
due to insufficient knowledge and reasoning capabilities,
bias and
lack of trustworthiness. Scaling generative AI
systems, particularly language models, has demonstrated
improved performance across a range of tasks, e.g., math word
problems and commonsense reasoning. However, even the most
powerful LLMs show a lack of robustness under different
semantic perturbations that would not have fooled an
otherwise robust reasoner
\citep{li-etal-2024-gsm}. While it is unwise to rule
out future improvements, their current limitations
call for interventions beyond scaling.
Thus, in order to be able to effectively solve complex
expert-domain problems, we believe it is necessary 
for the AI agent to work closely with human
experts.

Third, cooperative problem solving often involves the
\textbf{co-construction of the objective}
-- or \textbf{objective co-construction} --
alongside the
co-construction of the solution itself.
As part of the co-construction process, the requirements
for the solution are often changed and refined as the
collaborators understand the details of the complex problem better and
revise their initial assumptions. 
As we will see in
\secref{sec:framework}, we formalize this process of objective
co-construction through interactive preference learning:
we define a utility function over
the construction space that reflects preferences of user and
agent, i.e., which artifacts are better and which are worse
candidate solutions.
Thus, the objective is encoded in a
preference model. Or, stated differently, the objective is 
first described informally on a general level  --
e.g., write
a computer program performing a particular task -- and then formalized in terms of the
preference model.

One inspiration for making objective co-construction
part of our framework for solution co-construction comes
from the
field of 
multi-criteria decision aiding (MCDA),
a branch of operations research \cite{Roy1993,Roy1996}.
``MCDA underlines the 'aiding' 
in a process involving the DMs [decision makers] in the co-construction of their 
preferences \ldots\  It assumes that preferences of the DM
with respect to 
considered alternatives do not pre-exist in the DM's mind.''
\citep{Hullermeier2024part1}
MCDA (see also \citet{Hullermeier2024part2}) makes
assumptions that are close to objective co-construction; in
particular, the user's objective is only partly determined
in the beginning and further developed in the course of the
decision aiding process.

It is important to note that the two types of
co-construction in \framework\ -- solution co-construction
and objective co-construction -- are quite different.
Solutions are artifacts whereas objectives are encoded as
preference models.
The solution is co-constructed through search in the
construction space whereas
the objective is co-constructed through preference learning.
Still, at the highest level, both solutions and objectives
are the result of a cooperation of human and agent, with
each being a contributor 
even though their contributions  may
differ in nature and scope.

Fourth, in \framework,
\textbf{natural
language is the main communication medium}.
This
makes it
possible for human and AI agent to be \textbf{equal partners}, with
complementary strengths, in the co-construction process.
In our view,
language-based communication is 
a key enabler  of an equal partnership.
The ability to express oneself fully and on the same level 
is a prerequisite for making equal contributions to problem
solving.
Previous communication technology was a limiting factor in 
this regard: only with the advent of LLMs do we now have
AI agents available that comprehend
and generate natural language at a human level of capability. Such human-level capabilities
are required for the complex communication needs that occur
during cooperation on complex expert-domain problems.

There are certainly problem-solving scenarios in which
the human manages the process and the agent's role is
reduced to handling
low-level tasks (e.g., ``tool'' tasks like
internet search or copy-editing)
-- or, conversely, scenarios where the
human's role is reduced to
providing input when prompted by the agent (e.g., in active
learning). In contrast,
the
type of human-AI cooperation scenario we are
interested in is one in which the two partners are
equal.

Equality here does not mean identical roles.
On the contrary, 
the roles are complementary:
each partner has skills or knowledge
that the other lacks. 
An example for complementarity is that
the human may understand the context better in which the
complex problem arises (e.g., the requirements and
personalized preferences of human
stakeholders) whereas the AI agent may be able to more efficiently
access vast information resources and make more effective
use of tools such as compilers and unit testing.
Just as human-human cooperation excels at problem solving if the
collaborators complement each other, so is human-AI
cooperation most beneficial if each partner can contribute
their unique strengths.
\changetk{This aligns with recent evidence showing that human-AI cooperation alone does not guarantee superior performance; effective integration and task-appropriate division of labor are key to realizing the benefits of cooperation \citep{vaccaro2024when}.}

\begin{figure}[!t]
    \centering
    \includegraphics[width=\linewidth]{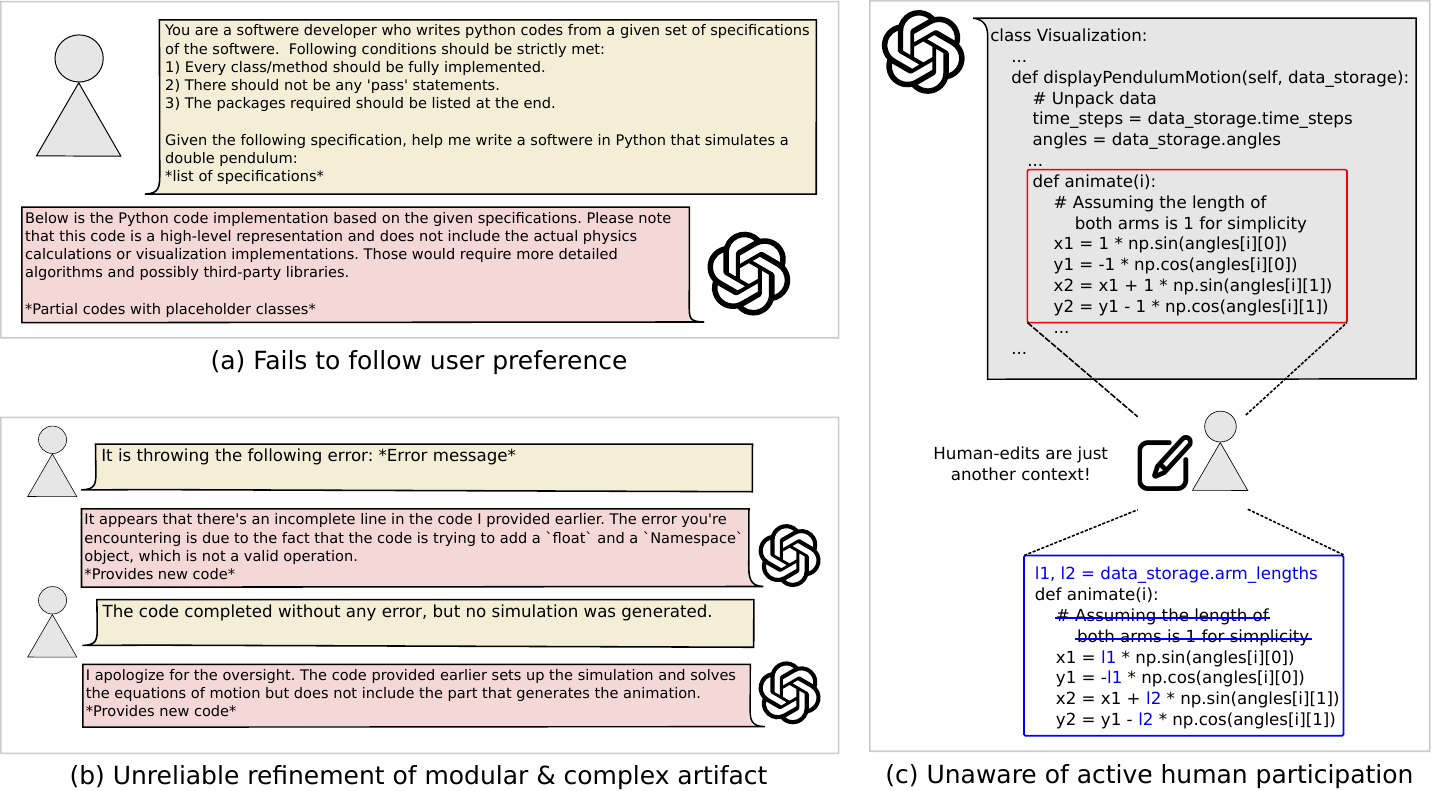}
    \caption{Existing generative AI lacks proficiency in key
      aspects of co-construction of solutions to complex
      problems.  We give a code synthesis example.  (a)
      GPT-4 Turbo fails to follow preferences explicitly
      stated by the human expert. (b) Due to the lack of a
      persistent object representation, a modification
      request targeted toward one feature of the desired
      solution leads to the unwanted (and erroneous)
      modification of another feature. (c) The human expert
      modifies the generated code directly to remove inline
      assumptions and introduces general variables; such
      active participation is not demarcated and recorded by
      the AI and there is no facility to extract the
      implicit preferences and follow them elsewhere. 
      }
    \label{fig:intro}
\end{figure}

\textbf{Scenarios that \framework does not address.}
To more clearly delineate which
class of scenarios we want
\framework to address, we now
give some examples 
where \framework is not a good fit for solving problems. 
(i) If the problem is simple (as opposed to a
complex expert-domain problem), then AI models probably can
solve it autonomously.
(ii) Even many complex problems may be solvable autonomously
by AI
(now or in the future)
if  a
full specification is available that can be checked
automatically. Games like Go and chess are
such examples in which objective co-construction is not
necessary, i.e., a full specification of the objective can
be easily obtained.
(iii) 
Another class of scenarios may be best addressed by an expert
managing the solution process and utilizing the agent for
solving specific subtasks. Somebody 
working through their email inbox after coming back from
vacation may want to closely oversee this process (e.g., not letting
the agent send email autonomously), but may be happy to
give specific subtasks to the agent (e.g., ask
correspondents that
requested reviews to  re-request them).

In contrast to scenarios
 in which the cooperation is dominated by
either the agent (ii) or the expert (iii), \framework is
intended for equal-partner scenarios in which the two work
together as partners leveraging their complementary strengths.



{\bf Is the current state of generative AI enough?}
As a running example, we will use a code generation
example that
illustrates some of the
bottlenecks of GPT-4 Turbo, an update\footnote{%
\changetk{We use \texttt{gpt-4-1106-preview}.}
} of the original GPT-4 model~\citep{openai2024gpt4technicalreport}. In Figure~\ref{fig:intro} (a), GPT-4 Turbo
ignores the human expert's explicit instructions to generate a
complete Python code with the required module
specifications, echoing \citet{xie2024travelplanner}'s
observation that current language-based AI agents lack task
adherence.  After repeated prompting with partial code
snippets, the process produces a complete -- albeit faulty --
code. This limitation is even more serious when
more varied and realistic expressions of human preferences are taken into
account -- for the human expert to contribute productively, one must allow preferences expressed via explicit instructions, binary choice, ranking, etc. Current generative AI solutions do not facilitate such multi-modal preference incorporation. Figure~\ref{fig:intro} (b) shows unreliable
debugging attempts. Specifically, the
LLM performs unrelated (and faulty) edits to address a bug
and
even introduces new errors. This demonstrates that existing
LLMs struggle with handling complex, modular software
code~\citep{jiang2024survey}.
The common practice is that
the human (as a knowledgeable 
expert who keeps track of overall context) identifies faulty output and
repeatedly prompts the model to guide it to the correct
generation -- this is implicitly adopting a co-construction paradigm.
However, Figure~\ref{fig:intro} (c)
shows that current modes of human-AI interaction cannot
unlock the full potential of co-construction -- direct
modification of the co-constructed candidate solution by the human
expert does not bear any special significance to the LLM, and
it treats it as just another context. There is no explicit
mechanism for the AI to learn implicit preferences expressed
by the human through active participation. 
{\color{black} On the other hand, forcing humans to take a passive role and making them review AI-generated code affects productivity negatively~\citep{code-review-challenge-I, code-review-challenge-II}. Meta-analyses by \citet{DBLP:journals/corr/abs-2402-11364}, with a focus on coding assistants, identify four major axes of AI-mediated productivity loss: shift of human roles from production to evaluation, unhelpful workflow restructuring, task interruptions, and, easy tasks becoming easier while hard tasks become harder. Recent advances in interactive coding have sought to address the first challenge by providing users with evaluation tools — AI-generated tests or static analysis-based — to decrease the cognitive load of reviewing AI-generated code. A key takeaway from these advancements is the need to redefine the human is supposed to do, what the AI needs to do, and how they are going to complement their respective expertise and limitations.
}

While these examples are focused on code synthesis, there is
evidence of similar shortcomings in other domains \changetk{
such as planning \citep{kambhampati2024llms},
radiology \citep{LECLER2023269}
and clinical decision making \citep{hager2024evaluation}.
}
{\color{black} \Citet{NEURIPS2019_f5b1b89d}, for example,
  demonstrate the necessity of incorporating explicit
  ``human awareness'' in a version of the collaborative game Overcooked, providing evidence that agents fail to coordinate with human subjects without such awareness.}
Code synthesis
in particular -- and the experience from day-to-day use of
generative AI for solving complex problems in general --
points toward co-construction as a naturally evolving
problem solving paradigm where the human expert tries to
search for the optimal solution by interacting with the
AI. However, the current state of generative AI hinders its
role as a reliable partner in successful co-construction.
This is because the ``one-directional'' interaction between human
and AI
typical of how AI agents are used today
often fails to steer the co-construction towards
a solution that satisfies the user's constraints.

In summary,
prior work has laid out the inherent shortcomings
of present-day generative AI
for complex problem solving (see also \secref{sec:rw} for a
much  more detailed discussion of prior work).
This motivates our alternative emphasis on human-AI
co-construction as a paradigm for solving complex problems
in expert domains.


\textbf{Ethical considerations.}
One aspect of \framework carries considerable risk:
the co-construction of the
objective. This opens the door to manipulation by the
agent. For example, LLMs that fail to solve a goal have been
observed to redefine it to be easier \cite{anthropic2025claude37}.

Influenceable reward functions have been studied by
\citet{10.5555/3692070.3692292}. They write:
``We show that despite its convenience, the static-preference
assumption may undermine the soundness of existing alignment
techniques, leading them to implicitly reward AI systems for
influencing user preferences in ways users may not truly
want.''
and:
``\ldots\  suggesting that a straightforward solution to the problems
of changing preferences may not exist.''

Even if there is no general solution to the problem of
influenceable reward functions, it is possible to  build
in safeguards in the context of \framework. Specifically,
if there is a conflict between user preferences and AI
preferences, then we can mandate that user preferences
prevail. This can be implemented by a ``monitor'' agent  -- a secondary
agent that cannot be manipulated by the primary agent and is
responsible for alerting the expert to objective changes
that were not clearly communicated. Alternatively, we can
employ the methodology of alignment to discourage
unwanted manipulation of the objective by the agent. For
example, we can devise a set of rules that should govern
objective co-construction (e.g., ``a change to the objective must
be clearly communicated to the human''), create a synthetic
dataset that embodies the rules and then train the agent on
this dataset using supervised finetuning or reinforcement learning.

More generally,
we believe that a paradigm of close cooperation is a
promising approach to addressing many of the hard ethical
problems that AI faces. If the agent takes an initial
problem statement, goes off and comes back with a complete
solution, then that means that the human cannot make any
course corrections.  This is true both for initial decisions
(if she were part of the co-construction process, the human
may realize that some initial decisions were based on wrong
assumptions and correct them) and for
decisions made by the agent (in which the human is not involved
in autonomous problem solving and therefore cannot influence).  Our co-construction
paradigm ensures consistent human participation in shaping
the solution as it is being constructed.  Similarly, if the
human is involved in the step-by-step co-construction of the
solution, then she will have a good understanding of its
inner workings and the motivation for its parts; thus,
cooperation can be effective in bringing about some measure
of explainability and (by extension) transparency with respect to ethical concerns.

In summary,
making the
objective influenceable by the agent is a risk.
But there are promising solutions for addressing it
(e.g., adding a monitoring agent for supervision).
In addition,
the close cooperation of human and agent
in \framework
-- which in contrast to an autonomous approach ensures that
the human is involved in all aspects of the co-construction process --
is a form of artificial intelligence that
addresses some ethics problems of AI by design, e.g., it
supports transparency with respect to ethical concerns.

{\bf Our contribution.}  In this article, we take steps
towards formalizing
co-constructive problem solving and thereby aim to address
important limitations of current generative AI models.
In contrast to approaches in which problems are solved
autonomously by AI or -- conversely -- in which AI is an
assistant without autonomy that is limited to executing tasks clearly defined
by the human, we view co-constructive problem solving as a
process that involves the two parties as equal partners, each
contributing complementary strengths.
Concretely,
we
present \framework{}, a conceptual framework 
that facilitates human-AI
co-construction. \framework introduces multiple levels of
abstraction to the candidate solution, providing a seamless
interface for the human expert and the AI agent to modify
and keep track of the complex, modular, co-constructed
candidate solution.  \framework{}
supports objective co-construction -- i.e., both the
solution and the objective are co-constructed by human and
agent --
by
allowing multi-modal
preference input from the human expert, with natural
language as the central mediator to capture information-rich
guidance signals, along with other forms of active expert
participation, such as categorical choice-based preference.
Solution co-construction in \framework{} is conceived as search
where the candidate solution (represented
on multiple levels of abstraction) is iteratively revised to
maximize the its utility, modeled by the preference
model. {\color{black}
While several components of \framework have been explored
in prior research independently across different domains
(see \secref{sec:rw}), we
are the first to bring them together under a unified
conceptual umbrella and to show \changetk{that,} enabled
by current generative AI, they have the potential to
address the major challenges in solving complex expert-domain
problems.}


\begin{figure}
\begin{center}
      \includegraphics[width=\linewidth]{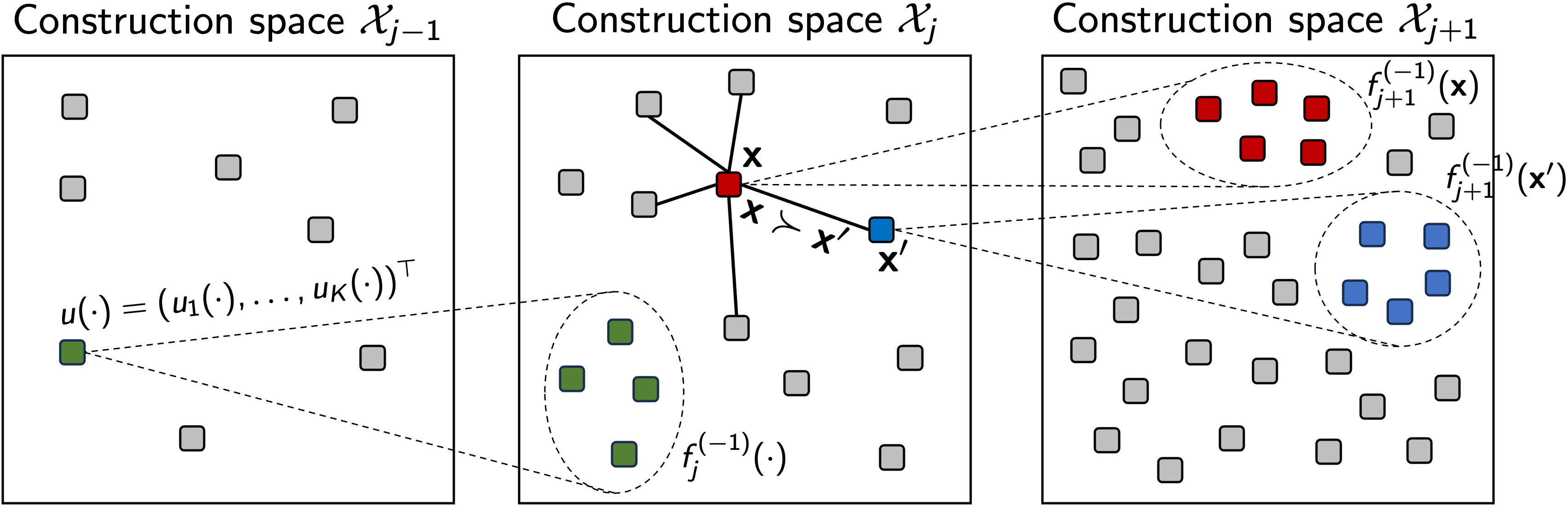}
\caption{Illustration of the hierarchy of construction
  spaces in \protect\framework. Each point $\vx$ symbolizes
  a candidate solution (on a certain level of abstraction),
  e.g., a software program. The topology of the space is
  specified by a suitable neighborhood structure (as
  illustrated for point $\vx$). Each point is associated
  with a latent utility $u^t$, possibly multi-dimensional and
  comprised of local utilities $u_1^t, \ldots , u_K^t$, and
  preferential information (e.g.,
 $\vx \succ \vx'$:
  solution $\vx$ is better
  than $\vx'$)
  that provides information about promising regions in the space. The relationship between the different abstraction levels is specified by the abstraction mappings $f_j$ resp.\ the (inverse) refinement mappings $f^{(-1)}_j$.
\label{fig:constructionspace}}
\end{center}
\end{figure}

\section{Towards a formalization of \framework{}}
\label{sec:framework}

In this section, we propose \framework, a framework for 
cooperative problem solving.
Broadly speaking, \framework is meant
to formalize an interactive problem solving scenario, in
which a human expert seeks to (co-)construct a \emph{solution}
 -- such as a computer program -- with the help of
 an AI agent.  The problem solving process is
conceived as a process of \emph{systematic search} in a
space $\mathcal{X}$ of \emph{candidate solutions}, i.e., as
a co-constructive process, in which candidate solutions
are modified or extended step by step until a (sufficiently)
good solution has been found. Therefore, we also refer to
the search space $\mathcal{X}$ as the \emph{construction
space}. The construction space, its hierarchical
organization and its topology (or neighborhood structure) are depicted in
Figure \ref{fig:constructionspace}.

Actions
taken by the AI agent
during the search
(e.g.,
adapting a candidate solution or
asking the expert a question regarding where to move
next)
depend on its \emph{informational
state} $\mathcal I$, which comprises its experience so far, e.g.,
about the expert's preferences, any relevant information
about the current context, the solutions considered so far and
the best solution constructed so far. Formally, the behavior of
the AI agent can be determined by a \emph{policy} $\pi$
that maps informational states to actions.

\begin{figure}[!t]
    \centering
    \includegraphics[width=\linewidth]{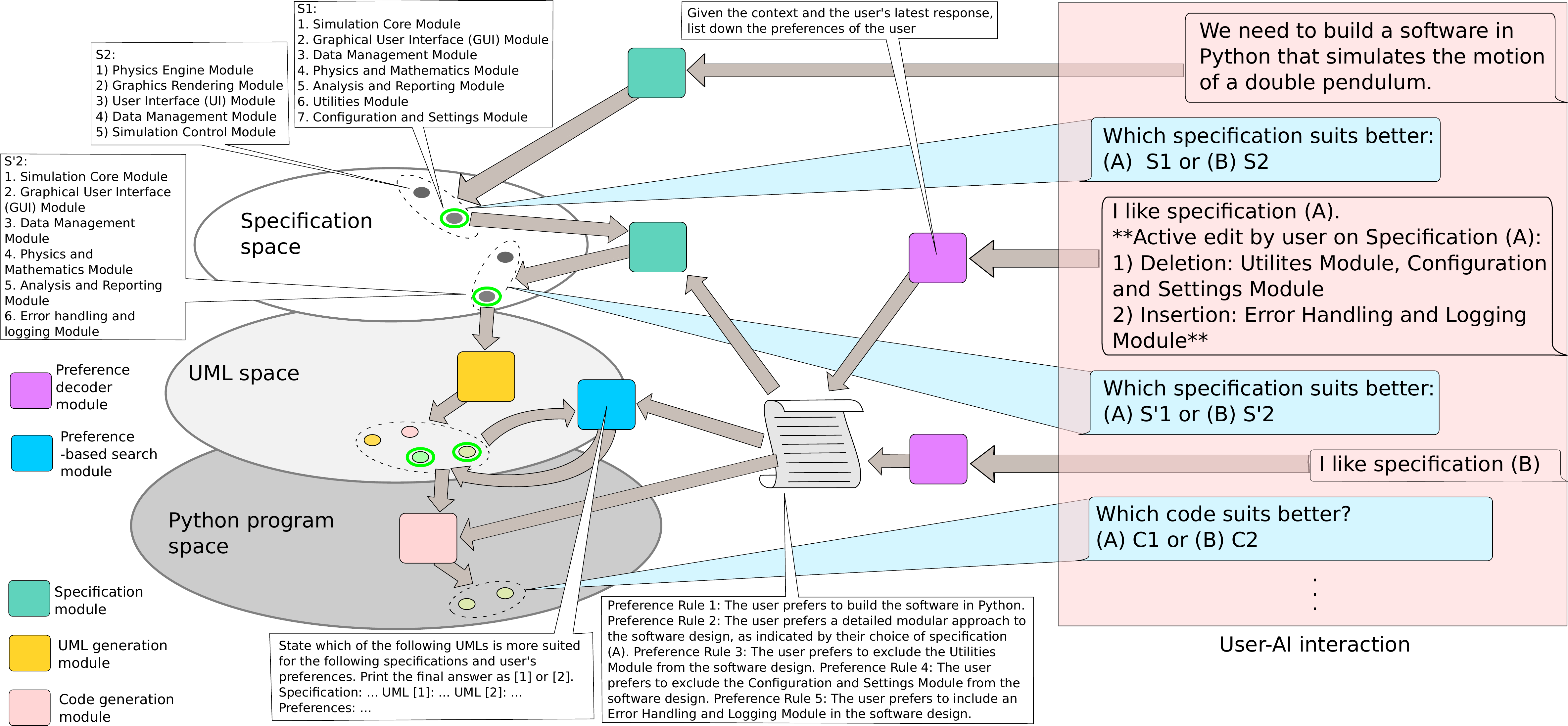}
    \caption{Running example. An example instantiation of \framework{} for the problem of building a double
pendulum simulation.  A solution is co-constructed 
through human-AI cooperation as follows. The interaction between the human and the AI is shown in the red box on the right, top to bottom. We define the co-construction space on three levels of hierarchy: Specification space, UML space, and Python program space. The user starts by specifying the (potentially incomplete) problem to solve. The Specification module (green rectangle) generates a pair of candidate specifications of the software to build (S1 and S2) and presents them to the user. The user expresses their preference in two manners: i) they choose one of the candidate solutions (S1) as better than the other, and ii) provide partial edits to the specification directly. The Preference decoder module (purple rectangle) extracts preference rules from the interaction context. Based on the decoded preferences, the Specification module generates a new pair of candidate specifications (only one shown for space reasons), from which the user chooses one. The UML generation module serves as a generator of refinements from specification space to UML space and generates a set of four UMLs from the specification selected. The Preference-based search module then runs a tournament-based search among these UML candidates: a pair of UMLs are compared against the specification and the decoded user preferences and one is chosen. Two finalist UMLs from the tournament are then used by the Code generation module (pink rectangle) to generate two candidate Python programs. These programs are presented to the user again for their feedback.}
    \label{fig:workflow-example}
\end{figure}

{\color{black}As a running example, we provide
  the code generation use case in
Figure~\ref{fig:workflow-example} as
  an illustration of how different formal aspects of
  \framework can be implemented  (see
  Appendix~\ref{sec:case-study} for more details).
  A second
  use case -- in the expert domain of related work section
  generation -- can be found  in Appendix~\ref{app:sec:rw-generate}.}
  {\color{black} The choice of code generation as a use case is motivated by recent advancements in automation in this domain and the subsequent development of knowledge regarding the limitations of these advancements. On the other hand, related work generation is a natural use case for the scientific community, irrespective of the individual researcher's particular domain.}

  {\color{black} \framework, as we envision, is not limited to these two choices of use cases. For example, in case of travel planning problems~\citep{DBLP:conf/icml/Xie0CZLTX024}, one can conceptualize an abstract hierarchy similar to multi-agent planning systems~\citep{DBLP:conf/iclr/LiXLTDL25} and a plan-space search~\citep{DBLP:journals/corr/abs-2501-09891} guided by human preferences. Objective co-construction under \framework\ allows the AI and the human to define the utility criteria of a plan together; e.g., an AI agent with access to large number of online reviews might inform the human about certain emergent patterns in hotel booking refusals, whereas an expert travel agent might guide the AI to book trips via certain transports. Similar analogies can be drawn for tasks pertaining to medical decision making. Recent studies~\citep{Tikhomirov2024-nc} have indicated inherent differences in the clinical reasoning process adopted by humans and AI systems. This further strengthens our argument for the goal of AI to become complementary to human, not mimicking humans. One can formalize the solution space as space of concrete diagnosis, with the abstractions highlighting different aspects of the diagnosis that construct a hierarchical plan~\cite{clinical-planning}.}
\subsection{Construction space and abstraction hierarchy}

The construction space will typically be large, most often even (countably) infinite. For example, the construction space may consist of all computer programs in a specific programming language. Spaces of this kind cannot be specified in an explicit way. Instead, they will be defined implicitly and may even be adapted or designed on-the-fly in the course of the problem solving process. In this regard, the \emph{formal representation} of candidate solutions is of major importance and will strongly influence the efficacy and efficiency of \framework. Moreover, it is also clear that the representation of solutions will not be universal but rather specific to the expert domain. For example, a computer program will not be represented in the same way as a machine learning pipeline or data science workflow. It should be noted that we do not make any assumption of {\em completeness} for candidate solutions: at any stage of the search, a candidate solution $\vx \in \gX$ can be partial or incomplete (i.e., an incomplete codebase, an incomplete ML pipeline, etc.). 

During problem solving, it is often useful to look at
(candidate) solutions on multiple levels of abstraction. In
many cases, for example, a rough draft of the solution is
found in a first phase of the process, and this draft is
then worked out in more detail in a second phase. More
generally, one can imagine a search process that switches
back and forth between different levels of abstraction
whenever appropriate. Therefore, we assume that the construction
space $\gX$ is equipped with a hierarchy of abstraction
levels. 

Formally, this can be modeled by a sequence $\gX_0,
\gX_1, \ldots , \gX_J$ of spaces, where $\gX_j$ is a
refinement of $\gX_{j-1}$ -- or, vice versa, $\gX_{j-1}$ an
abstraction of $\gX_j$.
We describe the abstraction process from $\mathcal{X}_{j}$
to $\mathcal{X}_{j - 1}$ as a surjection $f_j: \gX_{j}
\fromto \gX_{j-1}$ such that
$\vx'
= f_j(\vx) \in \gX_{j-1}$;
that is, $\vx'$ is the abstraction of $\vx$ on the abstraction level modeled by $\gX_{j-1}$. We
denote by $f^{(-1)}_j(\vx') = \{ \vx \in \gX_{j} \, \vert \,
f_j(\vx) = \vx' \}$ the set of all refinements of $\vx' \in
\gX_{j-1}$ on abstraction level $\gX_{j}$. Note that
refinements are not unique, which is why a transition from
$\gX_{j-1}$ to $\gX_j$ may come with a certain
arbitrariness. 

{\color{black} In our running example on code generation
  presented in Appendix~\ref{sec:case-study}, we implement
  the construction space on three levels of abstraction,
  considering a Python program as a refinement of a UML
  diagram, which in turn is a refinement of a natural
  language specification. The refinement maps are
  implemented by suitably prompted LLMs that take a
  representation in a higher abstraction space as input and
  produce a solution representation in the lower level of
  abstraction as output.  The one-to-many nature of the
  refinement is reflected via stochastic sampling in the LLM
  inference: multiple UML representations are sampled
  stochastically from one natural language specification. In
  case of related work section generation, these abstractions can be
  conceptualized as list of papers, semantic graph/outline
  of related work, and the actual related work section (the
  final artifact/solution, see
  Appendix~\ref{app:sec:rw-generate}). Note that in either
  case, the abstraction hierarchy can facilitate better
  utilization of different expertise of the human and the AI
  in the co-construction process; for example, modern code-language
  models are great at stylizing a piece of code (variable
  naming, commenting, etc.), while human understanding of
  requirement engineering is needed to reach the optimal
  specification document. Similarly, an LLM paired with a
  search engine can possibly result in an efficient
  paper-finder agent in the use case of related work
  generation, while the human scientist defines the utility
  function that evaluates a candidate related work section.}



 \subsection{Latent utility}
 
We assume that the construction space is equipped with a
latent \emph{utility function} reflecting the preferences of
the expert, i.e., the quality of solutions as perceived by
the expert.  But the utility function also reflects
input the AI agent may have given on which solutions are to
be preferred. Thus, the utility function embodies the
current state of the co-construction of the objective that
we introduced in \secref{sec:nonformalframework}. This also
means that the
utility function is not static, but changes as expert and
agent refine and change their goals during the
co-construction process.

What exactly do we mean by ``quality'' of the solution?
In general, ``quality'' may refer to various
dimensions or criteria, and different objectives might be
pursued at the same time; we formalize this with a
\emph{multidimensional} utility function $u^t(\vx) =
(u^t_1(\vx), \ldots , u^t_K(\vx))^\top$ comprised of local
utility functions $u^t_i$ where
the time index $t$ indicates the temporal dependency and dynamic nature of the utility function.
For example, a computer program
could be rated by average runtime or memory consumption.
The local utility functions can be combined into a scalar
utility function $U^t : \, \gX \rightarrow \mathbb{R}$ via a
suitable aggregation operator.

Various factors influencing the quality of candidate
solutions can be distinguished, notably hard and soft
constraints. \emph{Hard constraints} refer to (functional)
properties that qualify a candidate as a valid solution. For
example, a computer program should properly compile and not
contain any syntax errors. Even if invalid solutions should
normally be considered useless, the abstract notion of
utility is flexible enough to distinguish different levels
of invalidity. For example, a non-executable computer
program may still have a non-zero utility if the error can
easily be fixed by the expert. In any case, hard constraints will
normally not identify a solution uniquely. For
example, there are many computer programs that are
functionally equivalent in the sense of having the same
input-output behavior. \emph{Soft constraints} refer to
criteria that make a solution more or less desirable such as
the length of a computer program and its time and memory consumption.

In general, the utility (be it in the form of the multidimensional
utility function $u^t$ or the scalar utility function $U^t$)
is not known to the expert and the AI agent, nor are they explicitly
aware of it. 
Rather, the utility is latent  and underlies the
expert's preference feedback, potentially taking into
account the AI agent's input on what makes a good solution. 
Based on this, the AI
can 
learn an approximation $\hat{U}^t$.
The AI's goal is  then to construct a solution $\vx^*$ that maximizes $\hat{U}^t$, or which is at least close to the maximizer, while simultaneously improving the approximation quality of $\hat{U}^t$. 
The utility $U^t$ also induces utilities on higher levels of abstraction. For example, one way to ``lift'' a utility function from level $\gX_{j}$ to the more abstract level $\gX_{j-1}$ is via aggregation: $U^t(\vx') = \alpha ( \{ U^t(\vx) \, \vert \, \vx \in \gX_{j} ,  f_j(\vx) = \vx' \})$, where $\alpha$ is an appropriate aggregation function \citep{grab_af}. 

{\color{black}
In our running example, we represent the utilities as user preferences in natural language, extracted from the interaction where both the user and the AI agent can choose between presented options, provide explicit instructions, or actively edit parts of the solution. 
}


\subsection{Interaction and preference-based search}

Search through the construction space is guided by an
underlying \emph{search strategy} -- in principle, any
heuristic search method (properly balancing exploration of
the construction space and exploitation of acquired
knowledge) may serve as a point of departure. However,
in \framework,
the
search is also interactive and largely controlled by the
human-AI cooperation.



To guide the search,
human and AI can communicate via natural language; e.g.,
the AI agent may ask the expert for feedback or explicit
advice. Alternatively, the expert may actively intervene, for example
by critiquing or modifying a candidate solution. 
A third type of interaction, particularly important in the
context of \framework,
is driven
through \emph{preferential feedback}: By informing the AI
agent about the quality of candidate solutions, the expert
provides hints at presumably more promising (and, likewise,
less promising) regions of the construction space, and hence
suggests promising ``search directions'' to the AI
agent.
To give an example, the expert can compare two competing
candidate solutions with each other (e.g., whether a
modification has improved a solution or made it worse) and
provide this feedback (in natural language) to the AI agent
for the next iteration. The AI agent utilizes the feedback
to improve its approximation $\hat{U}^t$ of the latent utility
function, which is an important element of its informational
state.

{\color{black}
In our running example, for instance, we implement a preference-based search strategy 
that identifies promising solutions via a tournament of pairwise comparisons.
Besides, we realize a search policy that refines an existing
solution guided by the expert's preferences (see Appendix~\ref{sec:case-study} for details). Additionally, we present a conceptual application of \framework to another expert domain in Appendix~\ref{app:sec:rw-generate}: the task of related work generation~\citep{hu-wan-2014-rwgen, nishimura-etal-2024-rwgen}.
}

The way in which the AI agent and the human expert cooperate with each other is defined in the form of a \emph{protocol}. Among other things, the protocol clarifies the types of queries and responses on the two sides (AI agent and human expert) and the (preference) feedback that can be given by the expert.

In summary, the specification of a concrete \emph{instantiation of} \framework{} includes the following elements:
\begin{itemize}
\item[--] (Hierarchical) representation of candidate solutions (domain-specific)
\item[--] Structure of construction space $\mathcal{X}$, refinement/abstraction mappings, neighborhood structure
\item[--] Search operators (for modification of candidate solutions, refinement, abstraction, etc.)  and strategy 
\item[--] Natural language methods and protocol for cooperation
\item[--] Representation of informational states, the AI agent's action space and policy
\item[--] Utility as the formalization of the co-constructed
  objective: soft/hard constraints,  preference relations/predicates (i.e., what type of preferences can in general be expressed, and in which form)
\end{itemize}
While some of these components can be specified by hand, others could be subject to (machine) learning and data-driven adaptation.

\framework{} comprises a broad variety of human-AI
cooperation in natural language, as well as  categorical
choices and
active modification of the candidate solution by the
human expert. The search policy \(\pi\) is designed to
generate a (locally) optimal candidate solution based on the
immediate as well as historical feedback, thereby adapting
to the preference signals from the human expert user. The hierarchical abstraction of the search space facilitates a
modular modification of the complex candidate solution.
As we can see in the running example (Appendix \ref{sec:case-study}),
\framework{}
also allows for incorporating creative components into the
generation of candidate solutions, for example, through the
injection of randomness in the heuristic search process or the refinement of abstract into more concrete solutions.

\section{Challenges and future research}
\label{sec:challenges}

Our characterization of \framework implies multiple challenges that need to be addressed to realize co-construction effectively. In the following, we briefly describe these challenges with reference to the current state of research. {\color{black} Additionally, we outline possible future research along these directions. 

We start with the structural components of \framework in terms of the construction space and interaction between the AI system and the human expert.}

{\bf Specification of abstraction hierarchy.}
{\color{black}Two} core components of \framework{} are (i)
the abstraction hierarchy of the construction space
and (ii) the neighborhood structure that facilitates
preference-based search.
A synergistic
implementation of (i) and (ii)
poses a non-trivial research
challenge; several related criteria must be fulfilled, e.g., expressive and abstraction power of the hierarchical representations, ease of expressing human preference across different abstractions, aggregation of utility along the abstraction. In the domain of code generation,
\citet{le2024codechain} propose a modular approach to circumvent this challenge: they generate a
chain-of-thought style intermediate description of the
subtasks followed by modular codes implementing each of
them. Such a hierarchical generation approach can
be extended to solution
co-construction. However, relying on purely natural
language-based intermediate representations limits the
utility of the hierarchy -- structured, semi-symbolic
representations (e.g., UML diagrams for software) can
provide better abstraction and facilitate ease of
modification. {\color{black} Multi-agent systems adopting
  hierarchical workflows~\citep{DBLP:conf/acl/QianChatdev24,
    DBLP:journals/corr/abs-2409-HyperAgent} often simulate
  multi-level representation similar to human
  organizations. However, such abstractions are not
  formalized with the goal of co-construction. Such
  multi-agent systems suffer from limitations in
  specification and system design. We stress that future
  research toward designing the co-construction space needs to consider the strengths and limitations of the human expert and the AI system to best utilize the efforts of both. Simultaneously, the formal elements of the co-construction space (e.g., abstraction and refinement maps, utility aggregation, neighborhood structure) must be suitable for expressing expert preference.}

{\bf Communication in natural language.} When humans
co-construct a solution, communication in natural language
plays an important role. Natural language is a powerful and
at the same time succinct medium for conveying information.
Given the expressivity of natural language, human and AI
agent can easily communicate different options of how to
improve the current solution, both at a detailed level and
in more abstract terms~\citep{DBLP:conf/acl/QianChatdev24}. Similarly, preference learning is
facilitated by natural language, since many preferences are
easily specified in natural language. 
The challenge here is
that the language capabilities of LLMs have advanced to an
impressive level for the general domain, but this does not
apply to complex expert domains
\citep{magesh2024hallucinationfreeassessingreliabilityleading,hager2024evaluation,mezini24codellms}. {\color{black}It
  is essential for the AI agent to {\em understand} what the
  human partner (as well as other AI agents in case of a
  multi-agent setup) asks and describe what it needs from
  them, which remains an important open problem in present
  human-AI interaction
  research~\citep{bansal2024challengeshumanagentcommunication}. In
  this regard,
co-construction of
  objective and solution can serve as a point of fundamental rethinking: by formalizing a neighborhood and imposing a utility structure, it is possible to calibrate the effects of natural language-based interaction on the problem-solving process.}


{\color{black} Preferences of the human expert constitute the most important functional component of \framework. Next, we outline the challenges related to preference extraction in complex problem solving with co-construction.}

{\color{black}{\bf Learning from active human edits.} The
  role of the AI agent as a co-construction partner is
  central to \framework. This entails the possibility of
  active participation from the human expert and the need to
  learn expert preferences from such participation. Current
  generative AI lacks the necessary structures of the
  solution space, primarily represented in its input
  context, that could delineate the changes introduced by
  the expert and, subsequently, be the basis for learning
  from it. \framework{} provides a plausible alternative to
  ``put everything in context'' that can solve this
  challenge, as we argue in the following. Let $\vx\in
  \gX_j$ and $\vx'\in \gX_j$ be the solution before and
  after the edit from the human expert, respectively. The neighborhood
  structure imposed by \framework on $\gX_j$ requires
  learning the changes in the utility function upon moving
  from a candidate solution to a neighboring one. If one
  ensures a vector space structure formed by the utility $U^t(\vx)$,
  then the expert preference is equivalent to
  $U^t(\vx')-U^t(\vx)$.
  Even without learning to map the solutions to the utility, one can simply seek to learn the mapping from $\vx'-\vx$ to expert preferences under the assumption of a (locally) linear utility. \citet{gao2024useredit} previously showed that learning preferences from such changes is superior to prompting-based methods in terms of aligning LLMs to user edits. However, their experiments are focused on simpler, general-purpose natural language generation tasks. In expert-domain applications where the utility of a solution includes multiple hard constraints (e.g., executability of code) along with stylistic preferences, learning a structured representation of the utility space is essential and a challenge on its own.} 

{\bf Multimodal human-AI interaction.} Natural
language-based interaction is not the ideal channel for all
types of preferences.  Categorical preference can be
communicated more simply by pointing towards a better solution.  Thus, we would like to
incorporate multiple types of preference into \framework. {\color{black}Similar to deciphering the preference from natural language, different modalities and modeling approaches have their own sets of challenges and require non-trivial research efforts. For example, inferring a preference-based global ranking from pairwise comparisons can be challenging. Popular methods like the Bradley-Terry model~\citep{bradley-terry} have their own limitations, such as strong assumptions on the preference structure.
Prior literature tackling such hurdles~\citep{DBLP:conf/alt/MaoWR18,pmlr-v48-shahb16} paves the way for research under the umbrella of \framework.}
Additionally, incorporation of expert preferences across multiple modalities poses the challenge of {aligning} these multiple modes
of feedback with each other. For example, the human expert
may express the need for a security feature in a software
engineering problem explicitly, or they can express it
implicitly by choosing a candidate solution that includes
the feature over one that does not. The AI needs to extract
equivalent preference information in these two
scenarios. Contemporary research in recommender systems that
deal with modeling user preferences on multiple item
modalities~\citep{10.1145/3240508.3240541, 9152003} can
serve as a starting point. However, the relative complexity
and nuances of preferences in the case of \framework hinder a
trivial extension of recommendation-oriented solutions.

{%
{\bf Integration of vision-language models.}
The rapid advancement of vision-language models (VLMs)~\citep{%
li2024multimodal
} presents opportunities for extending how human experts communicate preferences within \framework beyond natural language.
While visual representations in construction spaces (e.g., UML diagrams) are already achievable through text-based specifications and external rendering, VLMs could expand this ability and also enhance how experts communicate preferences about solutions at any abstraction level.
VLMs would enable preference extraction from visual annotations, sketches, and spatial manipulations -- providing an additional channel for the expert to convey utility information, particularly useful in inherently visual domains such as UI development.
The key challenges mirror those already present in the text domain but with distinct complexities:
(i) defining neighborhood structures for visual construction spaces requires formalizing continuous visual operations into discrete semantic transformations, and
(ii) visual feedback signals must be decoded into the same utility learning framework that processes textual preferences, ensuring consistency across modalities.
Future research should investigate how \framework's mechanisms -- the informational state, preference decoder, and hierarchical consistency constraints -- can be extended to incorporate visual feedback while maintaining the systematic search methodology central to the framework.
}

{\bf Dynamic user preferences.} Current techniques of
aligning neural AI systems to human preferences, broadly
referred to as RLHF (Reinforcement Learning from Human
Feedback), typically involve a two-stage process: learning a
reward model on preference data followed by fine-tuning a
foundation model (often an LLM or a diffusion model) on
reward supervision from the reward
model~\citep{kaufmann2023survey}.  This setup is
fundamentally limited to static adaptation in the regime of
expert-domain co-construction; a single model of human
preferences is imitated by the agent that cannot adapt to the
personalized preferences of the human expert.  
This is a
fundamental challenge in co-construction problems, where the
AI agent must adapt to the evolving preferences of the human
expert.  Multi-turn RLHF~\citep{zhou2024archer}, although it
extends the context of preference-adherence to an iterative,
conversational regime, cannot solve the challenge of
dynamically evolving user preferences. The PAL framework, proposed by \citet{chen2024palpluralisticalignmentframework}, provides a partial solution to our problem via personalized modeling of static human preferences. Unlike traditional
policy learning, \framework{} motivates a reward-free
exploration of the solution
space~\citep{jin2020reward-free-explore}. In-context
reinforcement learning can pave the way towards handling
dynamic preference signals~\citep{yang2024rewardsincontext,
  lee2024ic-rl-1}. However, the action space in the scope of
\framework{} overlaps with the generation of multiple
hierarchical views of the candidate solution, rendering the
problem much harder than existing work on in-context
RL. Prior work with LLMs showcases the possibilities of
using them as in-context agents, though exploration
abilities will need fine-tuning-based
interventions~\citep{krishnamurthy2024largelanguagemodelsexplore}.


{\color{black}{
  {\bf Guardrails for objective co-construction.}
Co-construction of the
objective opens the door to manipulation to
AI agents 
\citep{NEURIPS2019_f5b1b89d,
  hong2023learning}; see 
\cite{anthropic2025claude37} for a real-world example. As we discussed in \secref{sec:nonformalframework}, even if 
there is no general solution to this problem,
we see promising avenues of research in the specific context
of \framework, including
introducing a secondary monitoring agent that alerts the human when
there is a suspicion of manipulative behavior by the primary
agent and employing the methodology
of AI alignment
for training the agent to refrain from
manipulative behavior -- a methodology that is an active
area of research with many challenges
\cite{casper2023openproblemsfundamentallimitations}, but has
nevertheless been successful in reducing the risks of
generative AI 
\cite{ji2025aialignmentcomprehensivesurvey,anthropic2025responsible}. Another perspective is to
broaden
the scope of observation. As
\citet{miehling2025agenticaineedssystems} argue in the
context of multi-agent systems, focusing on the individual
abilities of isolated agents in a multi-agent ecosystem can
lead to underestimation -- newer phenomena can emerge from
the inter-agent and agent-environment interactions. We
extend this argument to human-AI interactions and call for
agent integrity research under the broad vision of
co-construction
of solution and objective.
}

Finally, the search-based methodology outlined under \framework entails certain technical challenges inherent to the present-day AI, which we discuss in the following. Additionally, with the multi-turn feedback-driven process of problem solving, we discuss the non-triviality of evaluation as opposed to standard automatic evaluation techniques.}

{\bf Specification of informational state.} \framework
utilizes an informational state to keep track of
relevant information in the
interaction history. Given that the search policy is
conditioned on it, an efficient representation of the
informational state is a core challenge of
\framework. Typically, such interactive co-constructions are
expected to span a long sequence context. While we have
observed a significant surge in the context-size of
present-day generative AI (e.g., GPT-4 Turbo can handle up
to 128K tokens in the input prompt), recent research has
questioned the effective usability of such very long context
information~\citep{liu2024lost}. The representation of the
informational state needs to be compatible with the
abstraction specification of the construction space 
as well as the choice of how preference signals from the
human expert are encoded. This is particularly important as the
reflection of any preference signal upon the candidate
solution is manifested via the informational state -- an
unreliable update of the informational state subsequently
worsens the solution quality and may result in a divergent
search.





{\bf Creativity-correctness dilemma.}  The specific class of
co-construction problems that we seek to address requires
creative generation. At the same time, in most expert-domain
applications, the solution needs to fulfill 
objective correctness criteria. With generative models,
the two requirements of creativity and correctness become counteractive. Creative
generation typically emerges in highly stochastic regimes,
e.g., in high temperature
decoding \citep{wang2023costeffective}. However, increased
stochasticity carries the risk of
hallucination~\citep{aithal2024understanding-hallucination}. For problems with definite answers, it has already been
shown that more robust reasoning can be achieved by
stochastic exploration of the generation space and
identifying the subset of solutions that are most
consistent~\citep{wang2023selfconsistency}. However, such
self-consistency methods are limited to problem classes
with definitive answers  and cannot be readily applied
to the co-construction problems that we characterize in this
paper. In \framework{}, this can be generalized into a
broader learning problem of exploration-exploitation
trade-off. In the early iterations of co-construction, when
the preference input from the human expert is likely to be
vague, the AI may bias towards exploration of the
construction space in search for a creative solution
backbone. As the co-construction proceeds, the human expert
fixates on the feature requirements and the AI must refrain
from abrupt modifications and build on the preference model
developed from the early exploration.

{\bf Evaluation of co-construction techniques.}
Due to the dynamics of the
co-constructed objective
and the complexity and modularity of the solution,
the evaluation of co-construction is a difficult challenge.
We identify multiple dimensions of evaluation
that need to be addressed:
\begin{itemize}
    \item {\em Quality of the solution} should be evaluated
      using domain-specific measures; irrespective of the
      process of co-construction (of solution and objective), the solution must fulfill some objective criteria of correctness. 
    \item {\em Preference-adherence} is an essential
      criterion of the co-construction problem; across
      multi-iteration co-construction, the generation should
be compatible with the human expert's preference input.
    \item {\em Self-consistency} is another key aspect of \framework{}, as it allows multiple levels of abstraction along with multiple modes of human preference input; it is essential to quantize how consistently the hierarchical abstraction is represented and different modes of preference input are aligned.
    \item {\em Complexity} of co-construction includes the
      computational complexity of generation and
      preference-based search -- resulting in potentially high
      computational cost -- and the cognitive
complexity of the framework -- resulting in 
cognitive load for the expert user.
The latter demands significant research
      efforts from a multidisciplinary approach to ensure
      that automated assistants truly
      bring value to the expert.
\end{itemize}
Given that human experts are costly and have limited time,
LLM-based simulation of human-AI
interaction may facilitate large-scale
evaluation~\citep{tamoyan2024llmroleplaysimulatinghumanchatbot}. Even
though 
our four evaluation criteria seem to demand human evaluation, we conjecture
that the development of artificial critic
models~\citep{mcaleese2024llmcriticshelpcatch}, with human
value alignment, will be an important research direction in
the future.

\section{Related work}
\label{sec:rw}


We group relevant prior work into several major strands:
human-AI cooperation,
reinforcement learning from human feedback (RLHF),
assistance games,
learning from natural language interactions,
search- and evolution-driven construction,
inference-time compute scaling,
LLM agents for complex problem solving, 
persistent solution spaces for iterative construction
and human in the loop.
We discuss how these approaches attempt to address
(versions of)
expert-AI
co-construction.
However, as we saw in
\secref{sec:challenges}, they fail to comprehensively tackle these
challenges.



\changetk{%
\textbf{Human-AI Cooperation.}
Recent research emphasizes enhancing human-AI cooperation to support designers in complex, creative tasks.
A notable example is the AI-assisted design (AIAD) framework proposed by \citet{depeuter2023ai}, which mirrors our approach in emphasizing cooperation over automation and aiming to support designers' creativity by inferring their goals.
AIAD addresses the challenge of goal communication by using generative user models to understand designer reasoning and capabilities from their behavior.
This allows AI to provide helpful recommendations and learn from the designer's active participation and corrections.
More broadly, this line of work aligns with the vision of Hybrid Intelligence \citep{akata2020research}, which advocates for human-AI systems that interact as co-evolving collaborators with complementary skills and adaptive behaviors.
While both AIAD and Hybrid Intelligence share our goal of leveraging complementarity between human and machine, \framework{} differs from them in several aspects.
First, our approach explicitly adopts a hierarchical view encompassing multiple abstraction levels, like specifications, UML diagrams, and code, which is not a core element of AIAD or Hybrid Intelligence.
Second, the search in our framework operates across this multi-level structure rather than within a monolithic or unstructured design space, enabling systematic refinement and revision.
Third, we place stronger emphasis on actively learning not just from user feedback to AI suggestions, but also from direct human edits to evolving solution artifacts.
Fourth, \framework{} focuses on natural language communication for intuitive and flexible interaction between users and AI.
Finally, whereas AIAD emphasizes minimally disruptive assistance with the aim of AI unobtrusively supporting designers, our framework envisions a more proactive and equal partnership, giving the AI system more responsibility in the co-construction process, including proposing refinements to the objective.
}

\textbf{Reinforcement Learning from Human Feedback.}  RLHF
focuses on learning a policy preferred by humans, most
commonly relying on comparisons between candidate solutions
\citep{kaufmann2023survey}.  The goal is to learn a policy
that maximizes a reward or utility function that is
consistent with the human feedback.  Originating in classical
reinforcement learning domains such as games and continuous
control \citep{christiano2017deep}, RLHF has been
extended to a variety of domains, most notably fine-tuning
generative models
such as LLMs
\citep{stiennon2020learning,ouyang2022training}, eventually
leading to the development of AI models
 such as
ChatGPT
that can generate
human-preferred responses in natural language.

RLHF for generative AI is typically employed in a
single-turn setting, where the agent generates an immediate
response to a query, evaluated by a human expert.  This
contrasts with expert-AI co-construction, which involves
multi-turn interactions where agent and expert
collaboratively construct a solution.
Multi-turn interactions introduce challenges such
as extended time horizons and large action
spaces. Extensions to RLHF have been proposed
that address these issues
\citep{zhou2024archer}.

Even multi-turn RLHF, however, is not well suited to
expert-AI co-construction without further extension: It does
not maintain an explicit representation of the solution
space, which is crucial for systematic solution
construction.  In principle, RLHF could be used to learn the
AI agent's policy in \framework, but it is challenging to do
so interactively as required in \framework.

\textbf{Assistance Games.} Originally introduced as cooperative inverse reinforcement learning \citep{hadfield-menell2016cooperative}, assistance games \citep{shah2021benefits,laidlaw2024scalably} model human-AI interaction as a game under partial information where the AI strives to learn and maximize the human's underlying reward function.
A key assumption is that humans have pre-defined objectives, even if initially unknown to the AI.
While \framework{} shares this overarching aim of AI-driven assistance for human experts in tackling complex problems via iterative engagement and the integration of human preferences, it fundamentally differs in its formalization and underlying assumptions.

An important distinction is the concept that the objective itself is not fixed in \framework{} but co-constructed through interaction.
In complex domains, expert preferences rarely remain static; they evolve through exploration and iterative refinement. This dynamic formation of objectives stands in contrast to the assistance game paradigm, where the AI primarily infers a static, pre-existing human reward function. Beyond this conceptual distinction, assistance games offer elegant theoretical properties but face practical challenges in real-world deployment due to computational complexity arising from the many possible combinations of AI and human policies as well as human objectives.%
\footnote{\citet{laidlaw2024scalably} present steps toward scalably solving assistance games in more complex environments. While promising, our approach differs fundamentally through its emphasis on objective co-construction, structured search spaces, and natural-language communication.}
\framework{} addresses these challenges by structuring the search process at multiple abstraction levels where humans provide feedback at their most intuitive level, while leveraging pretrained language models to guide the search process.
While not suitable for all domains in which assistance games apply,
we argue that this combination of structured solution spaces, natural language communication, and emphasis on objective co-construction provides significant advantages in the domain of expert-AI cooperation.

{\color{black}\textbf{Learning from natural language interactions.} Natural language interaction between the expert and the AI agent is central to \framework{}. A popular approach towards facilitating human-AI interactive problem solving involves training the agent to follow instructions in natural language~\citep{branavan-etal-2009-reinforcement, tellex2011understanding}. This paradigm of learning to follow instructions has found attention in the LLM era as well~\citep{wei2022finetunedlanguagemodelszeroshot}. However, directly mapping language-specified goals to actions has limited applicability to novel tasks. Alternatively, learning rewards from natural language interaction to successfully align the AI agent with the human user has been explored~\citep{fu2018from,sumers2021learning} -- instead of learning to map language-defined goals to actions directly, they seek to learn the reward function from the language-defined goals that can be generalized to novel tasks. 
The findings of \citet{sumers-how-to-talk-to-ai} suggest that while instructions typically perform well in low-autonomy settings, high-autonomy regimes favor the reward-learning paradigm.
Instead of learning a language-conditioned policy or reward, it is also possible to use conversational cues as rewards themselves \citep{jaques-etal-2020-human}, which can be combined with the other approaches discussed here.
{\color{black}Yet another approach, deployed in OpenAI's
  ChatGPT, is to enable the language model to save
  information about the user and their interactions with the
  model functioning as a natural-language `memory', which may include information about the user's preferences.\footnote{\url{https://openai.com/index/memory-and-new-controls-for-chatgpt/}}}
Most of these approaches do not consider the need to actively elicit and co-construct the expert's preferences, a key aspect of \framework{}.
This necessity is supported by \citet{co-reyes2018metalearning} and \citet{lin-etal-2022-inferring}, who identify that inferring the correct behavior or reward from a single utterance is non-trivial given the multidimensionality of language.
Querying the human user and estimating their preferences in an interactive setup is also a key component of \citet{peng-PLGA}'s framework.
As a step in this direction, \citet{li2023elicitinghumanpreferenceslanguage}
use active elicitation to strengthen preference understanding -- the AI agent is trained to elicit and infer human preferences by actively interacting with the user.
This is crucial prior work for an implementation of \framework{}, forming an important pillar of future research under the abstract umbrella it provides.} 

\textbf{Search- and Evolution-Driven Construction.}  Our
framework emphasizes iterative search within the
construction space, a process akin to evolutionary
optimization, which iteratively generates and evaluates
candidate solutions \citep{back1996evolutionary}.  This
evolution can be viewed as a form of search-based
construction.
Interactive evolutionary computation, a
preference-based extension, is particularly relevant to our
work as it involves human evaluation of candidate solutions
\citep{takagi2001interactive,wang2024comprehensive}.
For
example, these methods have been applied to search-based
procedural content generation in video games
\citep{togelius2011searchbased}.
Our approach differs in
the core approach to the search process: Traditional
evolutionary methods maintain a population of candidate
solutions and generate new ones through mutation and
recombination.  In contrast, in our framework, each
iteration ends with a single candidate solution that is then
the basis for the next iteration. In addition, we leverage
the extensive prior knowledge of pretrained language models
to guide the search and use natural language communication
to facilitate cooperation between the AI agent and the human
expert.

{\color{black}
\textbf{Inference-time compute scaling.}
Allocating additional compute at inference-time can significantly enhance language models' performance on complex tasks.
Strategies vary along two axes: (1) \textbf{depth vs.~breadth} (sequential refinement vs.~independent exploration) and (2) \textbf{structured vs.~learned} process control (externally imposed rules and algorithms vs.~autonomous model capabilities guided by prompts or training).
Depth-based methods encompass \textit{structured refinement}, like systematic revision \citep{
   QU2024RecursiveIntrospection
},
and \textit{learned refinement}, such as autonomous reasoning steps via chain-of-thought \citep{
   cot,
   kojima2022large
}.
Breadth-based methods often involve structured parallel generation with verification (e.g., best-of-N, \citealp{
   cobbe2021training,
   lightman2024lets
}).
Recognizing complementary strengths of depth- and breadth-based approaches \citep{
  snell2024scalingllmtesttimecompute
}, recent work advocates combined approaches, either explicitly structuring integration \citep{
  wang2022selfconsistency,
  yao2023tree,
  snell2024scalingllmtesttimecompute
}
or relying on models' learned or emergent self-reflection and adaptive search-like behaviors \citep{
  openai2024openai,
  deepseek-ai2025deepseekr1
}.
Although both breadth-based parallel efforts and depth-based iterative revisions implicitly or explicitly search the solution space, this search is typically conducted non-interactively. This limits the ability to adapt generation to nuanced or shifting user goals \emph{during} the construction process, relying solely on pre-defined, static objectives or verifiers, with minimal incorporation of explicit human preferences or feedback during search.

In contrast to these predominantly non-interactive methods optimized for static objectives, our approach introduces an \textit{interactive, preference-guided search paradigm} explicitly designed to handle underspecified and evolving goals.
While learned reasoning techniques like chain-of-thought excel in domains with clear objective correctness, such as mathematical or symbolic reasoning \citep{sprague2025to},
our method targets complex expert tasks where subjective preferences and evolving requirements critically shape optimal solutions.
Instead of seeking solutions based on fixed objectives, we iteratively co-construct solutions aligned with dynamic expert preferences by explicitly integrating iterative feedback.
This allows the search process to dynamically reshape the model's reasoning trajectory in response to evolving goals, enabling alignment with expert decision-making in nuanced tasks.
Beyond this distinction, many of the approaches discussed above are complementary to our framework and could help produce better base models or be used as inspiration for the search process itself.

}

\textbf{LLM agents for complex problem solving.}
The rapid increase in the capabilities of  LLMs has triggered multiple recent efforts to integrate them at the core of autonomous agents that interact with the environment, plan, and act to solve complex problems~\citep{Wang_2024}.
Typical approaches adopt integrating different tool-usage capabilities into LLMs via efficient prompting, often with multimodal capabilities~\citep{chen2023endtoendembodieddecisionmaking}.
A single agent is often insufficient to solve complex
problems; thus, multiple agents with different capabilities
have to be integrated.
Recent efforts in LLM-based multi-agent systems seek to mimic such cooperative problem solving by role-playing LLMs via in-context examples~\citep{li2023camel} or fine-tuning~\citep{juneja2024textttlmtexttt2simplesocietylanguage}. {\color{black} Multi-agent systems have started gaining traction in expert domain applications as well, e.g., software development~\citep{DBLP:conf/acl/QianChatdev24}, finance~\citep{ganesh2024generativeFinance}, chemistry~\citep{MALLM-chem}. The ChatDev workflow proposed by \citet{DBLP:conf/acl/QianChatdev24} showcases the effectiveness of using natural language as the primary communication medium between roleplaying agents for software development. This demonstration in the context of AI-AI interaction strengthens our argument for natural language-based human-AI cooperation.}
These frameworks of LLM-based autonomous agents are largely designed toward AI autonomy. However, the target problems are not isolated applications -- they require interactions between human organization(s), and as a result, these autonomous agents end up as incomplete assistants. 
Recently, arguments in favor of strategically allocating
tasks between humans and LLM-based agents to exploit their
distinct strengths have been put
forward~\citep{he2024llmbasedmultiagentsystemssoftware},
which aligns with our approach of leveraging the strengths
of both humans and AI agents in
co-construction. {\color{black}We argue that the open
  challenges in efficient interaction between a human and an
  agentic ecosystem, as outlined by
  \citet{bansal2024challengeshumanagentcommunication}, share
  some of the characteristics of human-AI co-construction as
  we envision it.
An alternative view of our framework\footnote{Indeed, our
running example for \framework{} is closely analogous to multi-agent systems.} is hence an extension of LLM-based multi-agent systems with a human agent as a key component, focusing on the co-construction of solutions.}

\textbf{Persistent Solution Space for Iterative
  Construction.}  A fundamental component of our proposed
framework is the explicit representation of the construction
space for systematic solution search.
Such a persistent memory can be useful for LLM agents. 
\Citet{sumers2024cognitive} propose a cognitive architecture
for language agents that connects LLMs to internal memory
and external environments, grounding them in existing
knowledge or external observations.
Similarly, \Citet{modarressi2024memllmfinetuningllmsuse}
introduce a structured memory component that LLM agents can
use for storage and retrieval.
{\color{black}In terms of deployed products, Anthropic's artifacts\footnote{\url{https://support.anthropic.com/en/articles/9487310-what-are-artifacts-and-how-do-i-use-them}} add an explicit representation of an LLM-constructed artifact to the Claude series of language models, which can be iterated on through further interaction.}
Although these approaches
do not directly address expert-AI co-construction
challenges, they relate to our approach by providing agents
with persistent memory to store intermediate solutions and
relevant information for problem solving.

\textbf{Human in the Loop (HIL).}
In its original form,
HIL
refers to
a type of interactive machine learning system
in which the human has the role of an annotator, advisor or
provider of feedback that 
the system can solicit input from on specific
questions and requests \cite{mosqueira2023human}.
For example, in active learning, 
the machine learning system identifies
informative examples for the human to label and retrains the
machine learning model iteratively until a termination
criterion has been met.
While there is no generally accepted definition of HIL and
quite heterogeneous approaches have been grouped under this
umbrella term by different authors
\cite{McCarthy_Programs59,settles2009active,TOWELL1994119,schramowski2020,69f67f95f4554ddd9e164f4265e2f7f2,pub14233,10.5555/1619410.1619464,10.5555/777092.777132},
HIL work includes some form of human feedback
in contrast to completely automatic forms of machine learning.

In traditional HIL, the human's role is limited: it
is confined to the role of an oracle that is consulted with
specific requests. The human is not involved in higher-level
decisions or co-construction of a solution.

In contrast, as we discussed in \secref{sec:nonformalframework},
\framework is a framework in which human and agent
cooperate as equal partners, each fully engaged in 
co-constructing the solution to the problem and in
co-constructing
the objective.
Thus, the cooperation between human and agent is
more symmetric in \framework than in HIL.

The need for a more equal partnership between human and AI
agent -- going beyond a narrow definition of HIL -- is being recognized more generally.
For example,
\citet{natarajan2024humanintheloopaiintheloopautomatecollaborate}
argue that while there are problems in which the limited
role of humans typical of traditional HIL is appropriate,
many other problems (similar to the complex expert-domain problems
we target) require a more equal partnership.

{\color{black}
  \textbf{The term co-construction.}
We have borrowed the term co-construction from other disciplines, namely from the social sciences and humanities, where it usually refers to the ``joint creation of form, interpretation, stance, action,  activity, identity, [..] or other culturally meaningful reality'' \citep{Jacoby.Ochs.1995}. Although the definition allows for various possible interpretations, co-construction is interactional at its core. However, as the theory of co-construction is beyond the scope of this article and also out of our expertise, we would refer to \citet[Section~3]{Robertson.et.al.2024.BH} who thoroughly discuss constructivist principles and human-AI knowledge co-construction from a theoretical perspective. Beyond this theoretical basis, our scope is substantially different from \citet{Robertson.et.al.2024.BH} who solely consider efficient techniques for hand-crafting prompts for business managers.
}

\section{Conclusion}
\label{sec:conclusion}

Our
position is that existing generative AI agents require
active human participation to successfully construct
solutions to complex expert-domain problems, but cannot effectively serve
as reliable partners in human-AI cooperation due to their
current limitations.  We find evidence for this position in
prior research across a broad set of
domains. Our running example focuses on software generation
using GPT-4 Turbo, a strong proprietary LLM, and exemplifies
the major drawbacks of current LLMs such as inability to
follow human preferences, unreliable refinement of complex
solution artifacts and limitations to facilitate active
human participation. We observed that although day-to-day
usage of generative AI tends to adopt a type of human-AI
co-construction paradigm in an uninformed manner, the
challenges that LLMs face confine such interactions to a
much weaker form.

As a remedy, we introduce \framework{}, a
framework that is motivated by the effectiveness of
collective human problem solving. \framework{} facilitates a
solution construction space with multiple levels of
abstractions, in which human and AI iteratively refine the
candidate solution through search guided by human
preference. \framework{} allows active human participation
along with versatility in preference expression. After
presenting steps towards a formalization of \framework{}, we discussed the
research challenges -- including ethical challenges -- for this new approach as well as
possible future directions for addressing them.

\bibliography{tmlr,TK}
\bibliographystyle{compling}

\begin{appendix}

\section{An exemplary simulation of \framework}
\label{sec:case-study}


In this section, we present an example implementation of the major elements of \framework{},
tailored to code generation as a co-construction problem.
This example does not claim scientific rigor on its own;
instead, we use it to demonstrate what prior findings
(see \secref{sec:intro} and \secref{sec:rw})
already establish. {\color{black} 
We do not provide a complete implementation of  \framework; in particular, the following are not included in the case study: neighborhood structure of the solution space, preference extraction from actual human participation, dedicated utility function tailored toward the expert problem.  Instead, we emulate the intended behavior of a complete implementation using prompted LLMs, with the goal of motivating the practicality of \framework.}

The initial problem description is underspecified.  During
the cooperation, the user can introduce new requirements,
ask for modifications to the already generated code, and so
on.  We approximate different aspects of \framework{} 
(the surjective mappings between different abstraction
hierarchies of the construction space, policy and 
heuristic search strategy) using baseline implementation strategies for
ease of demonstration. Future research endeavors should be
directed to more in-depth implementation of these features.

{\bf Problem.} The user wants  to develop a
modular Python codebase for simulating a double
pendulum. Modules should include
components such as 
 I/O interface, visualization and physics engine.

In this example, the construction space \(\gX\) consists of the set of all Python programs. Three distinct levels of abstraction are implemented. (i) Specification space. A specification of the simulation software in natural language (\(\gX_0\)). (ii) UML space. A UML description of the software (\(\gX_1\)). (iii) Python program space. The Python program (\(\gX_2\)) itself. The abstraction refinements \(f^{(-1)}_1\) and \(f^{(-1)}_2\) (as introduced in \secref{sec:framework}) are implemented using suitably prompted instances of GPT-4 Turbo that we denote as {\em UML generation module} and {\em Code generation module}, respectively; while the former produces a (stochastic) set of refinements in UML given a natural language specification, the latter generates Python implementations of a given UML description. {\color{black}To decode the user's preferences from the interaction, we use a {\em Preference decoder module}, implemented using prompted GPT-4 Turbo. Following the focus on natural language, the informational state \(\gI\) is realized primarily as the interaction history in natural language, along with an explicit list of preference rules decoded from this interaction.
One can impose a geometric structure on $\gI$ by introducing explicit metric space structure on the different abstraction spaces (e.g., edit distance), rendering $\gI$ to behave like a trajectory. However, introduction of such structures will be dependent on the expert domain application.

To facilitate the exploration of the candidate
solution space, we generate multiple solution representations on different abstraction levels by setting a high decoding temperature
in the respective generation modules and sampling multiple
responses. Intuitively, we seek to exploit earlier findings
that a highly stochastic generation regime facilitates
better novelty~\citep{wang2023costeffective}. Furthermore, this imposed stochasticity can be interpreted as the notion of neighborhood in the respective spaces: one can treat two solutions sampled from the same input context of the generation module as neighbors; distance between two different input contexts can be measured by edit distance. Although we do not explicitly specify such geometric structure in this example, the search strategy uses it implicitly.}

We do not implement a concrete realization of the search policy \(\pi\); instead, we rely on the limited abilities of LLM instances to explore and implement policy iterations~\citep{krishnamurthy2024largelanguagemodelsexplore, brooks2023large}. {\color{black} While the notion of search is present across all three levels of abstraction, we perform explicit search in the UML space using the {\em Preference-based search module}, which runs a tournament among candidate UML solutions, guided by the decoded preferences.}


The implementation, as depicted in Figure~\ref{fig:workflow-example}, instantiates \framework{} as follows.
The co-construction starts with the user providing an underspecified description of the task (in this example, building a simulation software in Python). {\color{black}The specification module (green rectangle in Figure~\ref{fig:workflow-example}) generates the natural language abstraction of the candidate solution as a list of possible components of the software along with their functionalities. This serves as a transparent interface in natural language that provides a layout of the construction. A pair of candidate specifications are generated using a high-temperature stochastic generation regime. The user chooses one of them as better. Additionally, they can state any explicit modification request. Moreover, they can directly edit the specification if they have specific requirements in mind ({\em preemptive reviewer}), or choose to continue with the workflow and decide on the specifics upon observing the final program ({\em lazy reviewer}). The Preference decoder module (purple rectangle in Figure~\ref{fig:workflow-example}) lists down the preferences decoded from the user's actions. If the user introduces any new modification (e.g., in the presented example, they remove certain modules from the specification and insert new modules), the specification module generates a new pair of specifications for the user to provide feedback on. This continues till a suitable specification is obtained.} 

Next, the UML generation module generates a set of
stochastic refinements of the natural language specification
into UML descriptions. The UML description of
the software forces the subsequent code generation module to
generate a final program that consists of multiple,
independent components (in this case, Python classes) and
well-defined dependencies among such components. Micro-level
changes to the code (e.g., changing the design of the GUI,
choice of numerical algorithms in the simulator, etc.) can
be facilitated now without changing the complete codebase
-- a desirable property of our implementation that is closer
to real-life software engineering. This
addresses the challenge monolithic code LLMs face in
scenarios in which 
persistent editing is required.
However, generating the Python
programs from all such candidate UMLs and verifying them one
by one is both computationally expensive and infeasible for
the human user.

{\color{black} The preference-based search among the candidate UMLs is implemented as a tournament
by iteratively declaring one among a pair as the {winner} of a round.} After a logarithmic order of such
rounds ($\log_2 n$ being the depth of the tournament tree
for $n$ candidate UMLs), the Preference-based search module comes up with a final
pair and a summary of preference justifications.\footnote{See \url{https://subha0009.github.io/ExAIC-Interactions/PreferenceLayer.html} for the tournament on the candidate UMLs generated} Note that
although we seek to minimize the cost of human intervention
in this step by automating preference-based ranking, one
can envision the human expert providing their judgement on
these UMLs. In such a setup, the Preference-decoding module
can be used to explicitly adapt to such gold preference
examples.

Next, we utilize the code generation module to
translate each of the
two selected UML candidates into a candidate Python program
that will be used for human feedback post-execution. Aligned
to the goal of co-construction, in this last stage, the user
provides their binary judgment on the relative quality of
the two generated Python programs along with (optionally)
natural language feedback. Such feedback can incorporate the
errors found in the program execution (if any), additional
requirements, etc. This feedback, along with the summary of
the tournament generated by the preference learning module, are together used as a context for the next iteration of refinement. This iterative process continues until the user is satisfied with the solution.

{\bf Comparison with monolithic LLMs.} {\color{black} We perform offline human evaluation to compare \framework against a vanilla LLM (in this case, GPT-4 Turbo) in terms of their effectiveness as co-construction partners for expert domain problems. The problem to be solved is to generate code for the double pendulum simulation. Multiple co-construction episodes are generated by specifying different initial preferences and mid-episode preference switching (e.g., choosing different specifications generated by the Specification module; asking for different functionalities in the software; and editing different segments in the specification as well as the codes). Each human evaluator is presented with a pair of co-construction episodes -- one using \framework, one using GPT-4 Turbo -- and asked to compare them in terms of the following criteria:
\begin{itemize}
    \item[\bf Q1.] Which assistant has better incorporated the initial preferences of the user?
    \item[\bf Q2.] Which assistant better adapted to preference switching?
    \item[\bf Q3.] Which assistant is more precise in terms of iterative refinement? `Precision of modification' means 
    that the changes are relevant to the request.
    \item[\bf Q4.] Which assistant is more complete in terms of iterative refinement? `Completeness of modification' means 
    that all the necessary changes have been made.
    \item[\bf Q5.] Overall, which assistant seems more suitable for software-level code generation?
\end{itemize}

We recruited a total of 14 participants for this evaluation; each is a doctoral student in Natural Language Processing, so that expert preferences in code generation can be understood. We provide each evaluator with a pair of interactions between i) a user and GPT-4 Turbo~\footnote{Example interaction can be found at \url{https://subha0009.github.io/ExAIC-Interactions/Assistant2.html}} and ii) a user and \framework{}~\footnote{Example interaction can be found at \url{https://subha0009.github.io/ExAIC-Interactions/Assistant1.html}}. Figure~\ref{fig:enter-label} shows the interface used to collect the users' responses. We do not collect any personal information from the evaluators.

On all five criteria, the majority of evaluators rate \framework 
higher than GPT-4 Turbo.
 12 (85.7\%)  evaluators find that \framework better captures the initial preferences of the user ({\bf Q1}). 11 (78.6\%) agree that it can adapt better to preference switching  ({\bf Q2}). 
 11 (78.6\%) see \framework 
 as superior on 
 precision ({\bf Q3}) and completeness of modification
 ({\bf Q4}).
  12 (85.7\%) suggest \framework is better suited for software development ({\bf Q5}). Detailed responses are available here: \url{https://subha0009.github.io/ExAIC-Interactions/FormResponses.html}.

\begin{figure}
    \centering
    \includegraphics[width=0.5\linewidth]{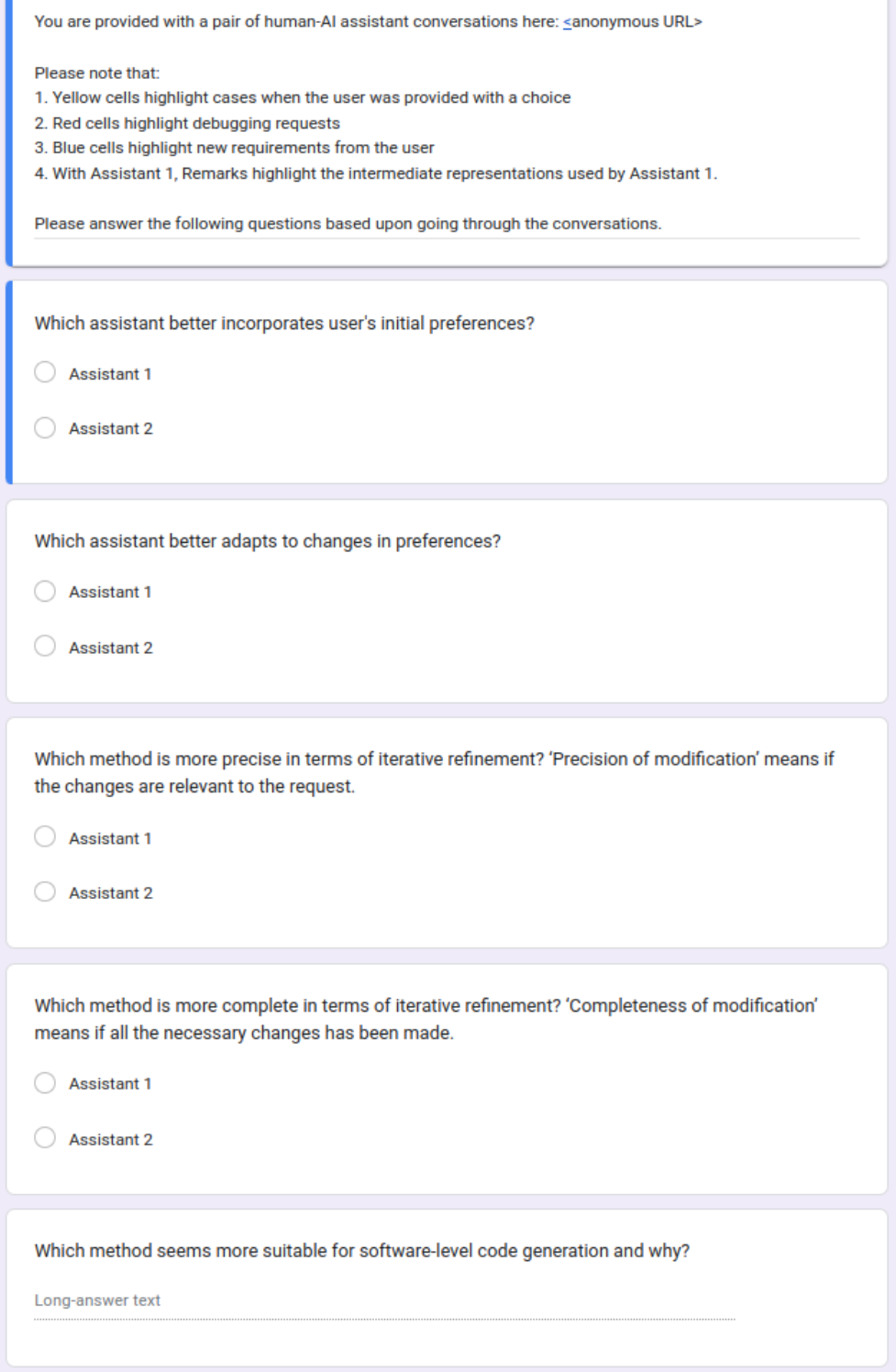}
    \caption{Interface used for human evaluation.}
    \label{fig:enter-label}
\end{figure}

{\bf Limitations and further improvements.}
The immediate improvements we observe are prevalent without
any dedicated implementation -- neither of the two
refinement maps
(from natural language to UML and from UML to Python) nor of
the preference-based search policy. In
this implementation, we do not equip the co-construction space with
explicit neighborhood structures. We posit that
development along these directions will further improve the
quality of co-construction and confirm \framework{}'s
potential as an effective framework for the class of
co-construction problems we aim to address. Modern software development
relies on software engineering tools such as type systems,
test drivers, static program analysis tools, monitoring and
debugging tools and security vulnerability detectors.
Realistic software artifacts are complex, and their full
evaluation by humans without these tools is infeasible.
Our implementation of \framework -- not intended as a systematic
evaluation of \framework's effectiveness in the software
domain -- will need to be extended with many of the elements
that are standard in the DevOps pipeline (see
\cite{le2022coderlmasteringcodegeneration,10.1145/3639476.3639770}
for examples
of how to integrate such standard tools). We use the
LLM's generative capacity as is, e.g.,  when it explains why
one generated solution is better than another.
An interesting research direction would be {\em explanatory interactive
learning}
\citep{ijcai2017p371,teso2019explanatory,friedrich_22typology},
where more faithful explanations are produced
through
interactively
constraining model explanations. {\color{black}The search strategy can be further refined by implementing reinforcement learning from execution feedback~\citep{gehring2024rlefgroundingcodellms, liu2023rltf, dutta2024frugal}.}

\section{Possible implementation in Related Work Generation}
\label{app:sec:rw-generate}
\begin{figure}[!t]
    \centering
    \includegraphics[width=\linewidth]{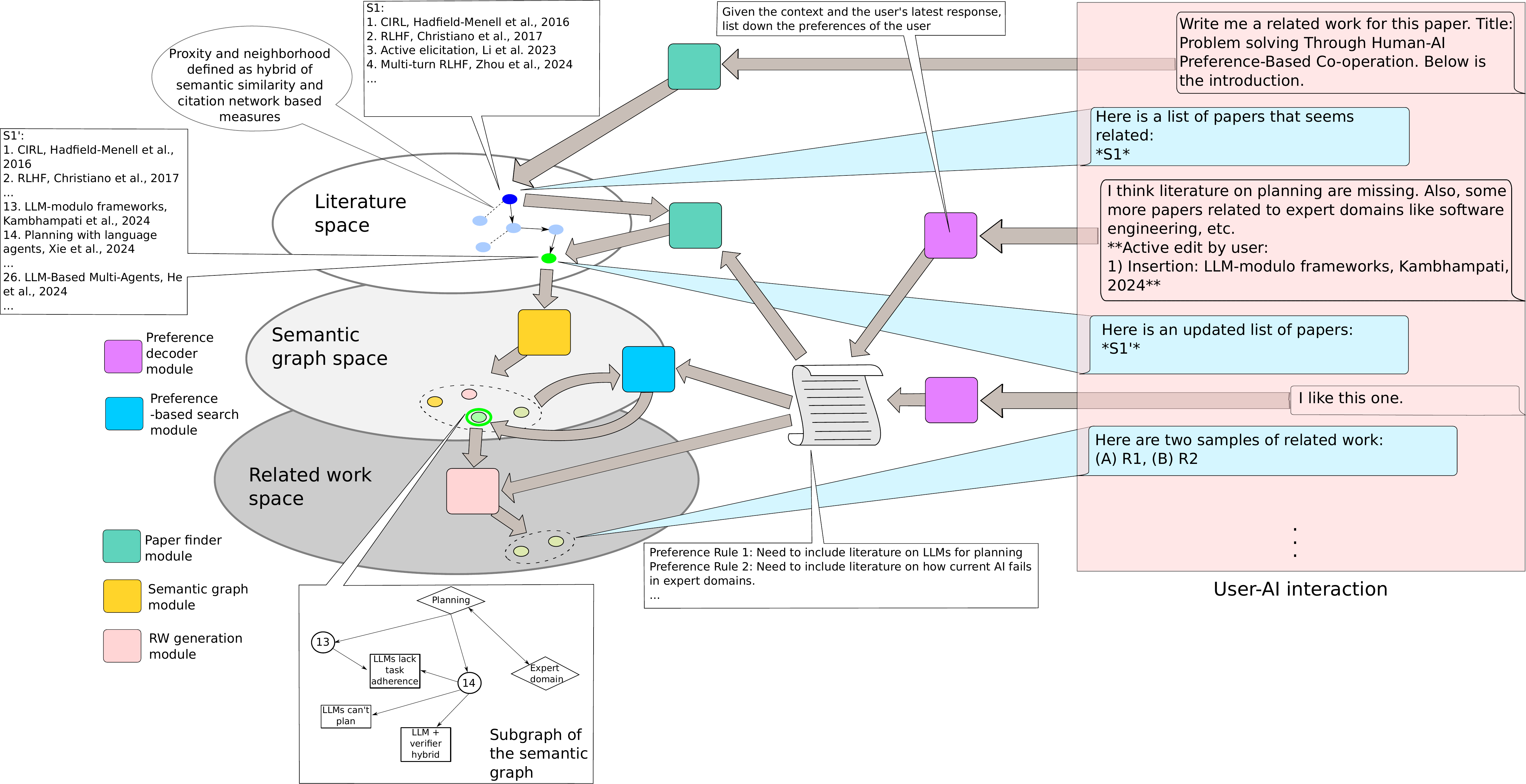}
    \caption{\color{black}A conceptual application of \framework for the problem of generation of a related work section of a research paper. The user provides the title and the introduction of the research paper for which the related work section is to be co-constructed (i.e., written). Three abstraction hierarchies are envisioned. The Literature space consists of lists of papers. The Semantic graph space depicts the papers, their findings, and their interrelations using a directed graph. 
    The Related work space contains related work sections (written in natural language
    using a writing style appropriate for this genre)
    that describe the semantic graph. The Paper finder module lists down the  papers that should be incorporated into the related work. Neighborhood structure is imposed upon this space using semantic similarity and hops over the citation network. The Preference decoding module keeps track of the user's preferences. Upon deciding on a suitable list of papers, the Semantic graph module translates them into a semantic graph. Finally, the Related Work (RW) generation module writes related work sections based on the semantic graph.}
    \label{fig:related-work-generation}
\end{figure}
To showcase the applicability of \framework for expert domains other than software engineering, we present a  workflow for 
the problem of 
related work generation in Figure~\ref{fig:related-work-generation}. Note that this is not an actual implementation, rather a proposal on how \framework can be adapted for this problem. Similar to our case study on software engineering, a solution construction space with three levels of hierarchy is defined. The highest level of abstraction (Literature space) is the space of lists of relevant papers; each point 
(represented as a list of papers)
is intended to capture the literature relevant to the research paper (in this case ``Problem solving through Human-AI Preference-Based Co-operation''). Based on the user's problem specification, the Paper finder module (which can be an LLM-web search hybrid) lists the possible papers that are relevant to the research paper. This is a classical search problem. Next, these papers are used to construct a semantic graph that relates different papers according to their domain of focus, methodology, findings, prescriptions, etc. Such a graph is inherently heterogeneous. Multiple semantic graphs can be generated from a given list of papers. One can define a neighborhood over the space of these semantic graphs via edit distance. The search strategy, in this case, can again be realized through a
tournament (as in the software engineering domain presented in Section~\ref{sec:case-study}) or through another mechanism
(e.g., a specialized module that evaluates semantic graphs based on their
graph-theoretical properties). Finally, the RW (related work) generation module translates these semantic graphs into Related work sections. Locally valid neighborhood structures can be constructed using the neural representations of textual differences. The Preference decoder module extracts the preferences expressed by the user to guide the search in different spaces. In this example, one can define refinement maps between the different abstraction levels in a straightforward manner: The semantic graph can be mapped to the list of papers directly as the former has nodes that are members of the latter. Similarly, each pair of nodes and their connecting edge in the semantic graph can be translated to a sentence in the related work section in the Related work space.

}

\end{appendix}

\end{document}